\documentclass[11pt]{article}
\usepackage{amsmath}
\usepackage{amsfonts}
\usepackage{amssymb}
\usepackage{amsthm}
\usepackage{graphicx}
\usepackage{tabularx}
\usepackage[table]{xcolor}
\usepackage{hyperref}
\usepackage{float}
\usepackage[font=small,labelfont=bf,justification=justified,singlelinecheck=false]%
{caption}
\usepackage{fullpage}
\usepackage{natbib}
\setcitestyle{authoryear, round} 
\usepackage{subcaption}

\title{Representation Learning on a Random Lattice}
\author{Aryeh Brill\thanks{Email: aryeh.brill@gmail.com. This work is supported by a grant from the Long-Term Future Fund (EA Funds).}}
\date{October 31, 2024; Revised March 13, 2025}

\begin{document}
\maketitle

\begin{abstract}
    Decomposing a deep neural network's learned representations into interpretable features could greatly enhance its safety and reliability. To better understand features, we adopt a geometric perspective, viewing them as a learned coordinate system for mapping an embedded data distribution. We motivate a model of a generic data distribution as a random lattice and analyze its properties using percolation theory. Learned features are categorized into context, component, and surface features. The model is qualitatively consistent with recent findings in mechanistic interpretability and suggests directions for future research.
\end{abstract}

\section{Representation Learning and Target Functions}

To make sense of low-level sensory data, one can construct a representation of its underlying explanatory factors of variation. In machine learning, good performance usually relies on a representation that draws out and disentangles important information \citep{bengio2013representation}. Deep learning's enormous success on many real-world tasks arises in large part from its ability to learn useful representations from data without relying on built-in human knowledge \citep{sutton2019bitter}. 

An ideal representation disentangles explanatory factors into independent, atomic features that are either sparse or composable, making the model more robust to task and distribution shifts \citep{bengio2013representation}. Decomposing a neural network's representations (i.e., activations) into independent features is a fundamental objective of mechanistic interpretability research \citep{bereska2024mechanistic}. In particular, sparse dictionary learning using sparse autoencoders (SAEs) has received significant attention as a method to extract interpretable sparse latent features from a neural network’s activations   \citep{bricken2023monosemanticity, cunningham2023sparse, templeton2024scaling, gao2024scaling, lieberum2024gemma}. In general, the ability to identify, interpret, and audit the features used by neural networks would be a crucial step toward ensuring these systems' safety.

Unfortunately, ``feature'' is an elusive concept. Mechanistic interpretability researchers have explored several working definitions \citep{elhage2022superposition}. First, a feature could be any arbitrary function of the input, but this seems too broad. Intuitively, a feature should provide a uniquely useful abstraction. A narrower definition is a property that corresponds to a human-understandable concept. While this subjective definition is appealing for interpretability research, there's no guarantee that a model's internal operations align with human concepts. Finally, \citet{elhage2022superposition} defined a feature as any property of the input such that there should exist a sufficiently large neural network that would represent that property using a dedicated neuron. A downside of this definition is that it's circular -- features are assumed to be useful without explaining why.

An illuminating perspective on features comes from the geometry of the data distribution. Suppose the input data are embedded in a high-dimensional space $\mathcal{X} \subseteq \mathbb{R}^d$. Empirically or by assumption, valid real-world inputs may be confined near one or more low-dimensional regions $\mathcal{M} \subset \mathcal{X}$. Commonly, $\mathcal{M}$ is thought to approximate an embedded (nonlinear) manifold, with $\dim\mathcal{M} < \dim\mathcal{X}$. This \textit{manifold hypothesis} is widespread in machine learning \citep{scholkopf1998nonlinear, cayton2005algorithms, van2008visualizing, bengio2013representation}. The manifold hypothesis suggests that good features correspond to intrinsic coordinates on $\mathcal{M}$.

Alternatively, a data distribution may be better modeled as a set of discrete, disconnected clusters. In this case, a good representation might consist of a categorical distribution over prototype vectors or cluster centers \citep{linde1980algorithm, lloyd1982least, van2017neural}.

A data distribution is a set of valid inputs. What makes an input valid? Partly, it's an objective fact about the external environment's data-generating process. Only some conceivable world states are consistent with physical laws and the world's contingent circumstances. Deep neural networks trained on disparate objectives, datasets, and modalities may develop convergent representations due to this shared reality \citep{huh2024platonic}. But the model or agent's subjectivity also plays a role, in two ways\footnote{A similar interplay between objective reality and subjective observation arises when formulating macrostates in statistical mechanics \citep{shalizi2003macrostate}.}.
First, the input format reflects many choices, including sensor modality and resolution; dataset composition and preprocessing; inductive biases of the model and training procedure\footnotemark; and built-in task-specific knowledge. These factors influence the data distribution's embedded geometry, even if constrained by objective, extrinsic structure. Second, the task to be done determines which input information is useful. Well-constructed features just need to map the projection of the data distribution that's relevant, filtering out disallowed or unimportant sources of variation. Good representations are goal-dependent.

\footnotetext{For example, natural images possess symmetries such as locality and translation and scale invariance. Accommodating these using convolutional architectures \citep{lecun1998gradient} or data augmentations \citep[e.g.][]{chen2020simple} generally improves performance. In this work, we consider such inductive biases part of low-level input processing, with our focus instead on representations of high-level semantic content. Intuitively, at scale, learning problems are dominated by the world's unbounded contingent complexity, not fixed perceptual symmetries. This unboundedness reflects the principle that ``the environment is larger than the agent'' \citep{soares2015, demski2019embedded}.}

A task to be done can be thought of mathematically as a target function $f: \mathcal{X} \to Y$, where $Y$ is some output space. In machine learning, $f$ is typically implicit and learned from data. Imposing $f$ on $\mathcal{X}$ yields a data distribution. The geometrical properties of the data distribution under a given target function provide natural constraints on the representation used by an optimized model.

The target function can have different interpretations in different machine learning settings. For supervised learning with a single task, there's a clear target function which produces a narrow system using a specialized representation. Alternatively, unsupervised learning on a diverse corpus can train a more general system. Large language models (LLMs) pretrained on next-token prediction of internet data fall into this category. In this case, the target function would seem to map out the full data distribution, producing an all-purpose representation that can be specialized for particular tasks via prompting, finetuning, or distillation. 

An all-purpose representation must be very large, potentially leading to inefficient downstream computation \citep{chalmers1992high}. A generally intelligent system may benefit from having flexible representations for different frames and levels of abstraction. Some proposals envisage a system with a high-level controller that can configure its perception and behavior for the task at hand \citep[e.g.][]{chollet2019measure, lecun2022path}. In this case, the target function can be understood as the desired system behavior when configured for a specific task.

\section{Random Lattice Model of a Data Distribution}

\citet{brill2024neural} presents a random lattice model of data distributions found in natural learning tasks, which this section reviews. While that work's primary focus is predicting neural scaling laws, our aim is instead to investigate the model's implied data distributional geometry. 

\subsection{Motivation}

We model the data space $\mathcal{X}$ as a $d$-dimensional hypercubic lattice. This choice assumes some finite discretization, perhaps related to a practical data resolution limit. The target function is defined by input examples $x \in \mathcal{X}$ with corresponding targets $y \in Y$. For convenience, the output space $Y$ is also assumed to be discretized. We next make two key assumptions.

\textit{1) Context-dependent target function:} Many realistic target functions can be viewed as an accumulation of disparate behaviors. An LLM might write Python code, compose poetry, and play chess. We model this property by decomposing the target function into a higher-order function $HOF$ that yields first-order functions $f_x: \mathcal{X} \to Y$, such that $HOF(x) = f_x$ and $f_x(x) = y$. Next, we connect an adjacent pair of input examples $x_i$ and $x_j$ with a bond if $f_{x_i}(x_i) = f_{x_j}(x_i)$ and $f_{x_i}(x_j) = f_{x_j}(x_j)$. That is, we connect elements of $\mathcal{X}$ where the target function behaves the same way. Chains of connected elements comprise clusters. Elements with bonds are in-distribution or occupied, because their functions can be learned by generalization. Elements with no bonds are out-of-distribution, because their functions must be memorized. 

\textit{2) General-purpose learning:} The learning system contains no built-in knowledge about the data or task to learn. Its input format is statistically unrelated to the data distribution's extrinsic latent structure, which derives from the endlessly complex outside world \citep{sutton2019bitter}. Because the input format is arbitrary, we model its statistical properties by assuming that it's random. 

Let $p$ be the fraction of occupied sites. Our data distribution model consists of a $d$-dimensional lattice with sites occupied at random with occupation probability $p$. The input dimension $d$ can be assumed to be large for realistic datasets.

\subsection{Percolation Theory Analysis}

\begin{figure}
    \centering
    \includegraphics[width=\linewidth]{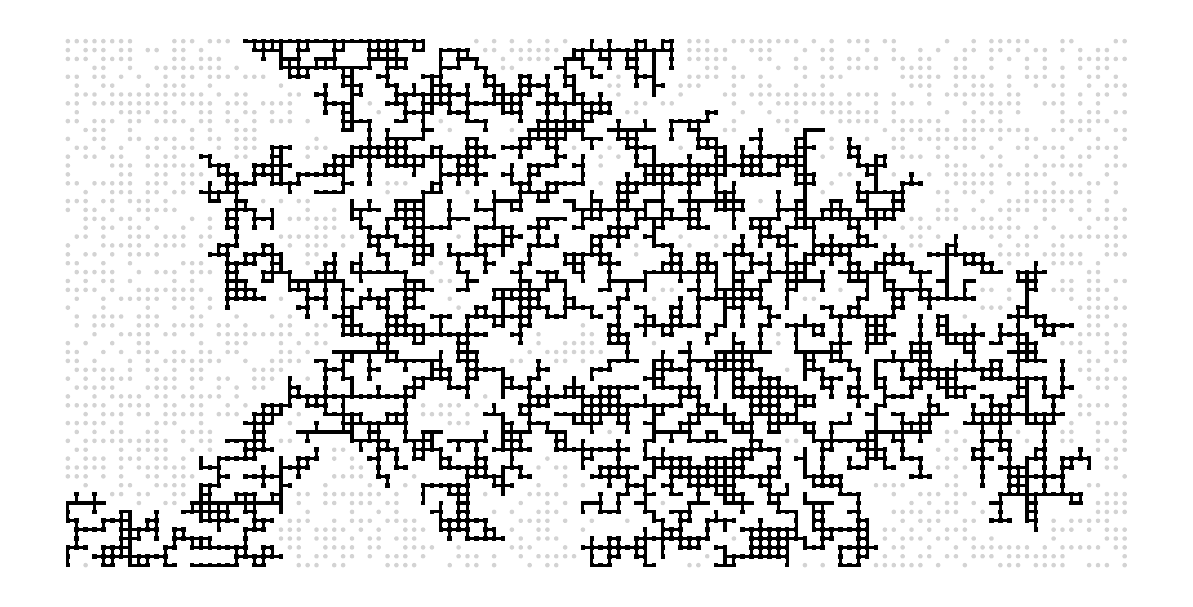}
    \caption{Percolation at $p_c$ on a $60 \times 120$ square lattice. Connected black sites show the largest cluster.}
    \label{fig:percolation_illustration}
\end{figure}

A hypercubic lattice of randomly occupied sites can be analyzed using percolation theory \citep{stauffer1994introduction}\footnotemark. Adjacent occupied sites form clusters, and percolation theory describes these clusters' statistical and geometrical properties. The essential finding is a critical phase transition: above a critical occupation probability $p_c$, the system is dominated by a single ``infinite'' cluster that \textit{percolates} or scales with the system size, while below $p_c$, only finite clusters exist. Fig.~\ref{fig:percolation_illustration} illustrates two-dimensional percolation at the critical threshold.

\footnotetext{Site percolation analyzes randomly occupied lattice sites, while bond percolation analyzes randomly occupied bonds between sites. The distinction is immaterial in high dimensions, and we elide it. Following \citet{stauffer1994introduction}, we use the language of site percolation.}

At $p_c$, finite clusters have a power-law size distribution for large $s$, $n_s \sim s^{-\tau}$, with the cluster number $n_s$ defined as the number of clusters of size $s$ per lattice site. Away from $p_c$, the power law has an exponential cutoff scaling with $|p - p_c|$. Above $p_c$, the infinite cluster contains most occupied sites, with finite clusters making only a subdominant contribution. For all $d \ge 6$, $\tau = 5/2$.

Percolation clusters have fractal geometry. For a given $d$, all finite clusters have the same fractal dimension $D < d$, where $D = 4$ for all $d \ge 6$. The incipient infinite cluster at $p_c$ is also a fractal object, exhibiting self-similar geometry at all length scales. For $p > p_c$, the infinite cluster's intrinsic dimension increases, and it becomes Euclidean when $p \gg p_c$. In low-dimensional percolation, clusters contain ``blobs'' of multiply connected sites, as Fig.~\ref{fig:percolation_illustration} shows. However, in high dimensions, $d \ge 6$, self-loops are vanishingly rare and clusters look treelike. Appendix~\ref{appendix:percolation} reviews derivations of some of the above properties for percolation in high dimensions.

These results provide a quantitative model of data distributional geometry for data distributions near the percolation threshold. In the subcritical regime where $p \lesssim p_c$, data distributions consist of disconnected finite clusters with small fractal dimension. The clusters' power-law size distribution corresponds to a power-law distribution of usefulness for prediction. Each cluster supports an effectively distinct target function or ``quantum'' \citep{michaud2024quantization}. In the supercritical regime where $p > p_c$, the data distribution is dominated by the infinite cluster, corresponding to an embedded data manifold.

\section{Features from Data Distributional Geometry}

\subsection{Three Categories of Natural Features}

The percolation model suggests three categories of natural features to be learned. These features are a minimal set of sparse or composable latent variables that map the data distribution, allowing the target function to be computed efficiently. First, \textit{context features} hierarchically identify the clusters and subclusters to which inputs belong. Second, \textit{component features} provide a coordinate system on each cluster so its function can be computed. Third, if general-purpose learning is imperfectly realized, \textit{surface features} represent predictive structure in the input's form.

Context features map each input to its corresponding cluster (for $p \lesssim p_c$), and potentially to subclusters as well. Large clusters (including the infinite cluster for $p \gtrsim p_c$) have fractal substructure. The substructure's self-similarity suggests that subclusters may themselves have subclusters, potentially down to the dataset's ultimate level of resolution. Thus, there may be a hierarchy of low-level context features mapping to subclusters at increasingly fine levels of granularity, in addition to high-level context features mapping to fully distinct clusters.

In fact, mechanistic interpretability studies have shown that language models learn a diversity of context features \citep{bills2023language, gurnee2023finding, bricken2023monosemanticity}. For example, \citet{gurnee2023finding} identified multiple apparently monosemantic high-level context features learned by the Pythia suite of LLMs \citep{biderman2023pythia}, including neurons associated with French text, Go code, and US Patent Office documents. In addition, \citet{bricken2023monosemanticity} employed SAEs to extract many interpretable features from one-layer transformers. Notably, \citet{bricken2023monosemanticity} found that increasing the autoencoder size often produced ``feature splitting'', in which a coarse feature at small size splits into a set of related fine-grained features at large size. Feature splitting may be consistent with models learning a hierarchy of multi-level context features, in turn reflecting cluster substructure in the data distribution.

Next, component features provide the coordinates for computing each (sub)cluster's function after its lowest level of identifiable substructure is reached. Under the assumed context-dependent target function, every input example $x$ in a cluster must have approximately the same first-order function $f_x$, which can be modeled as a single cluster function. A (sub)cluster's component features are compositional variables corresponding to the primary axes of variation under this function. Equivalently, the object mapped by a set of related component features could be usefully described as one multidimensional feature \citep{engels2024not, olah2024linear}. Recently, \citet{engels2024not} conducted a theoretical and empirical investigation of multidimensional features in language models, finding evidence for multidimensional representations of days of the week and months of the year.

Finally, surface features are the result of any departure from our definition of an ideal general-purpose learning system. At scale, these should be an exceptional minority of features. For example, a language model might improve performance by exploiting surface-level statistical correlations and parroting memorized associations. An example of surface features in image models may be the ``non-robust'' features exploiting superficial correlations in the data that can give rise to adversarial examples \citep{ilyas2019adversarial, gilmer2019a}. Another class of surface features are those related to mathematical operations, such as modular addition \citep{liu2022towards, nanda2023progress, zhong2024clock, engels2024not}. These representations are determined by logical necessity, not empirical reality, making the general-purpose learning assumption inapplicable. In addition, real models deal with the data format to process data and make predictions. Computations related to this may yield surface features.

\subsection{Emergence of Latent Features}

A power-law distribution of useful latent features naturally emerges from our model. Features have a geometric interpretation as a coordinate system for mapping the data distribution. Furthermore, features are ultimately defined functionally, embodied by clusters of inputs on which the target function behaves similarly. 

Clusters are self-similar objects. As the model size and the dataset's size and resolution increase, the number of resolvable latent features grows in principle without limit. A given finite model learns only a finite number of features. However, with increasing scale, models can identify more clusters and resolve finer substructure in known ones, causing context features to be added and component features to be added or modified. Clusters' hierarchical substructure means that the emergent latent feature space might be productively thought of as a tree, rather than as a set.

Clusters, and features, can be understood in different ways depending on the setting. In a supervised setting, the clusters identified by context features might correspond to subtasks. Alternatively, we could think of them as concepts that describe some relevant property of the world. This may be especially appropriate in a self-supervised setting in which the task is modeling the data distribution itself. Viewed in this way, concepts hierarchically decompose into related, specialized sub-concepts, modeled as self-similar cluster substructure. Concepts are sparse in the subcritical regime; almost all are irrelevant to any given input. Far less sparsity is expected in the supercritical regime, occurring only for low-level context features resolving the infinite cluster's substructure. When a concept is relevant, its corresponding component features determine the appropriate functional behavior. Finally, surface features correspond to concepts that have essentially logical rather than empirical content, or have no semantic content at all.

Our percolation analysis yielded discrete, unambiguously separable clusters because the dataset space is assumed to be discretized. Adjacent dataset-space elements are either connected or they're not. However, this binarization is best understood not as fundamental, but as approximating a putative underlying context-dependent target function. A richer approximation might allow bonds between elements of dataset space to have varying weights. Such a model may better describe more realistic concepts defined by inexact family resemblances, in which elements can have more or less typical membership \citep{Wittgenstein1953, rosch1975family, yudkowsky2008}.

\subsection{Identifying Learned Features}

Each feature category has a predictable pattern of sparsity and density. In general, context features should be highly sparse when viewed in the appropriate basis, since each input example belongs to only one cluster. Different clusters' component features should exhibit mutually sparse activations, for the same reason. This sparsity suggests that models may efficiently represent these features polysemantically, in superposition \cite{elhage2022superposition}\footnote{Particularly important high-level context features may be represented monosemantically \citep{gurnee2023finding}.}.

On the other hand, a model might use many low-level context features at varying levels of abstraction to map out a single cluster. It then would be normal for multiple such context features to activate simultaneously. Furthermore, the component features belonging to a given cluster act in composition. In both cases, models might naturally employ dense, compositional representations.

The expected relative proportions and absolute numbers of each feature category depends on the dataset. Table~\ref{tab:feature_regimes} provides a heuristic summary. The occupation probability $p$ parameterizes dataset regimes\footnote{Conceptually, a small value of $p$ would indicate a more context-dependent target function. Further work is needed to formalize and quantify this relation.}. In the subcritical regime where $p \lesssim p_c$, the data distribution consists of many distinct clusters with a power-law size distribution, requiring many context and component features to describe. In the mildly supercritical regime where $p \gtrsim p_c$, the infinite cluster dominates the data distribution. As a low-dimensional fractal object, it presumably can be mapped using relatively few context and component features. In the extreme supercritical regime where $p \gg p_c$, the infinite cluster is a high-dimensional object with Euclidean geometry. Many component features, but no context features, are needed to describe it. 

\begin{table}[htb]
    \centering
    \begin{tabular}{c|c|c}
        Regime & Context Features & Component Features \\
        \hline
        $p \lesssim p_c$ & many & many\\
        $p \gtrsim p_c$ & few & few\\
        $p \gg p_c$ & none & many
    \end{tabular}
    \caption{Predicted feature prevalence by category in different dataset regimes.}
    \label{tab:feature_regimes}
\end{table}

\subsection{Application to Interpretability: Sparse Autoencoders}

An exciting development in mechanistic interpretability has been the development of SAEs, which have been used to extract seemingly interpretable features from LLMs \citep{cunningham2023sparse, bricken2023monosemanticity, templeton2024scaling, gao2024scaling, lieberum2024gemma}. An SAE is an autoencoder that decomposes a dense vector into an overcomplete sparse basis using a loss function that regularizes for sparsity in addition to minimizing reconstruction error. However, SAE feature reconstructions appear unsatisfactory in several ways, calling into question the method's underlying inductive biases. SAE reconstructions appear to have nonnegligible irreducible error \citep{gao2024scaling}, with the errors having predictable structure \citep{engels2024decomposing}. Although SAE features are supposed to be independent and atomic, instead feature visualization \citep{bricken2023monosemanticity, templeton2024scaling} and clustering \citep{engels2024not} methods reveal groupings of semantically related features. \citet{li2024geometry} discovered complex geometric structure in SAE feature space at multiple scales. Furthermore, feature splitting \citep{bricken2023monosemanticity} appears to indicate that SAE features are often not atomic, but instead have decomposable structure \citep{leask2025sparse, chanin2024absorption}.

In light of these findings, an active research area now is to develop interpretability methods that reflect the true structure of features better than conventional SAEs do \citep{mendel2024, hindupur2025projecting}. Because a trained network's latent features should recapitulate the data distribution's geometry, our model may inform potential directions for interpretability research.

A fundamental assumption underlying any sparse dictionary learning method, including SAEs, is the superposition hypothesis \citep{elhage2022superposition}, which states that the latent features to be learned are sparsely activating, and stored in a nearly orthogonal overcomplete basis. For an SAE or similar method to be useful, it's essential that the underlying data distribution engenders sparsely activating features, which then could be stored in superposition. However, efficient alternatives such as composition, in which latent features activate densely, also exist \citep{olah2023distributed}. Our model predicts the preponderance of feature sparsity and density depending on the dataset regime. Table~\ref{tab:feature_regimes} implies that sparse dictionary learning can be mainly expected to be useful for datasets in the subcritical regime, where feature sparsity predominates. Even in that regime, not all features are sparse. Latent context features at multiple levels of abstraction may activate in composition, as would all component features for a given cluster.

Another essential assumption for SAEs, broadly influential in mechanistic interpretability, is the linear representation hypothesis \citep{bricken2023monosemanticity, park2023linear}. Informally, it states that features (1) occupy one-dimensional subspaces (directions) in activation space, and (2) behave mathematically linearly. Our model predicts the existence of multidimensional features for representing component features, contrary to hypothesis (1), but in line with the theoretical and empirical investigations of \citet{engels2024not}. Specifically, component features should generically require four-dimensional representations, as they map out treelike data volumes that have four-dimensional fractal geometry. This picture contrasts with the circular feature manifolds discovered by \citet{engels2024not}. Context features do not necessarily need multidimensional representations, so they may be consistent with hypothesis (1). Because our model is architecture-independent and instead based on data distributional geometry, it does not address the validity of hypothesis (2). It may be a fruitful research direction to explore interpretability methods sensitive to finding multidimensional features, perhaps by clustering sparse SAE features \citep{engels2024not} or by more directly extracting mutually sparse sets of dense latent features.

An SAE attempts to extract features as a sparse set. If this inductive bias were appropriate, one might expect that scaling up an SAE would simply recover more true features until they're all found. In reality, while larger SAEs do recover some entirely novel features \citep{leask2025sparse}, they also suffer from feature splitting \citep{bricken2023monosemanticity}. In particular, larger SAEs can split off low-level special cases from an otherwise interpretable high-level feature \citep[absorption,][]{chanin2024absorption}, or learn fine-grained features composing more fundamental high-level concepts \citep[composition,][]{leask2025sparse}. These findings suggest that SAEs may be imperfectly recovering true features that have a nested hierarchy. That picture is consistent with our model's predictions. First, both context and component features should have a power-law distribution in use frequency, providing a natural ordering for larger SAEs to learn novel features. Furthermore, percolation clusters' statistical self-similarity suggests that they could be naturally modeled by a hierarchy of increasingly fine-grained context features. Component features and low-level context features might be expected to activate conditionally on higher-level context features. An interpretability method for discovering features that map this fractal data geometry should have an inductive bias for hierarchical structure. Recent efforts towards developing SAE variants using nested groups of latents \citep{nabeshima2024, bussmann2024} may be a promising step in this direction.

\bibliography{main}

\appendix

\section{Review of Percolation Cluster Properties}\label{appendix:percolation}

This section reviews selected properties of percolation clusters, applicable when $d \ge 6$, $s \gg 1$, and $p \approx p_c$. We derive properties of the clusters' size distribution, including the power-law exponent and exponential decay away from criticality, and their fractal dimension. In particular, we approximate percolation on a high-dimensional lattice using the Bethe lattice, a special system that corresponds to an infinite-dimensional lattice and admits exact solutions. For more details, the reader is referred to \citet{stauffer1994introduction}, \citet{bunde2012fractals}, and references therein.

The Bethe lattice is an infinite connected undirected graph with no cycles (a tree) in which all nodes have equal degree $z$. The lattice centers on an arbitrary origin site, which connects to $z$ neighbor sites. Each neighbor further connects to $z - 1$ new sites, with each of those sites in turn connecting to $z - 1$ new sites, such that the branching continues indefinitely\footnote{If branching continues only for a fixed number of iterations, the resulting finite structure, called a Cayley tree, has nonnegligible boundary conditions \citep{ostilli2012cayley}.}. The Bethe lattice's $l$th shell therefore contains 1 site if $l =0$ or $z(z - 1)^{l -1}$ sites if $l \ge 1$. We consider site percolation, in which each lattice site is randomly occupied independently with probability $p$. If the Bethe lattice is used to model a $d$-dimensional hypercubic lattice, then $z = 2d$.

We begin by examining structural properties of percolation clusters on the Bethe lattice. To do so, we define the correlation function $g(l)$ as the expected number of sites at a distance $l$ from any occupied site that belong to the same cluster. Here $l$ represents the path length, also called chemical distance. Any two sites on the Bethe lattice belonging to the same cluster must be connected by a complete chain of occupied sites. We therefore have, for $l \ge 1$,

\begin{equation}\label{eq:corr_fn_l}
    g(l) = z(z - 1)^{l - 1}p^l = \frac{z}{z - 1}\left[(z - 1)p\right]^l.
\end{equation}

By inspection, $g(l)$ goes to zero exponentially with distance if $p < 1/(z - 1)$ and diverges if $p > 1/(z - 1)$. The critical percolation threshold is therefore

\begin{equation}\label{eq:p_c}
    p_c = \frac{1}{z - 1}.
\end{equation}

Next, we define the correlation length $\xi_l$ as a measure of the average distance between any two sites belonging to the same cluster,

\begin{equation}\label{eq:corr_fn_l_def}
    \xi_l^2 = \frac{\sum_{l=1}^\infty l^2 g(l)}{\sum_{l=1}^\infty g(l)}.
\end{equation}

The sums can be computed using Eq.~\ref{eq:corr_fn_l},

\begin{equation*}
    \begin{split}
        &\sum_{l=1}^\infty g(l) = (1 + p_c)\sum_{l=1}^\infty x^l = \frac{(1 + p_c)p}{p_c - p},\\
        &\sum_{l=1}^\infty l^2 g(l) = (1 + p_c)\left(x\frac{\mathrm{d}}{\mathrm{d}x}\right)^2\sum_{l=1}^\infty x^l = \frac{(1 + p_c)p}{p_c - p}\frac{p_c(p + p_c)}{(p_c - p)^2},
    \end{split}
\end{equation*}

\noindent where we defined $x \equiv (z - 1)p = p/p_c$ and used $z/(z - 1) = 1 + p_c$, yielding

\begin{equation}\label{eq:corr_len_l}
    \xi_l^2 = \frac{p_c(p + p_c)}{(p_c - p)^2} \propto (p_c - p)^{-2},~p < p_c.
\end{equation}

We can also straightforwardly compute the mean size $S$ of finite clusters by summing over the expected number of connected sites at all distances,

\begin{equation}\label{eq:mean_s_l}
    S = 1 + \sum_{l=1}^\infty g(l) = \frac{p_c(1 + p)}{p_c - p} \propto (p_c - p)^{-1},~p < p_c.
\end{equation}

We next consider the size distribution of finite clusters. We define the normalized cluster number $n_s(p)$ as the expected number of clusters of size $s$ per lattice site, for occupation probability $p$. In percolation, sites are occupied independently at random. The number of empty neighbors that surround a cluster is called its perimeter $t$. It follows that

\begin{equation}\label{eq:n_s}
    n_s(p) = \sum_t g_{st} p^s (1 -p)^t,
\end{equation}

\noindent where $g_{st}$ counts the possible cluster configurations with size $s$ and perimeter $t$. For the Bethe lattice, Eq.~\ref{eq:n_s} can be simplified because $s$ and $t$ have an exact relation. A single site has perimeter $z$ and adding a site to a cluster increases its perimeter by $z - 2$. Thus $t = (z - 2)s + 2$, and

\begin{equation}
    n_s(p) = g_s p^s (1 -p)^{2 + (z - 2)s}.
\end{equation}

The number of cluster configurations $g_s$ is difficult to calculate. It can be eliminated by considering the ratio $n_s(p)/n_s(p_c)$. Keeping terms of lowest order in $(p_c - p)$,

\begin{equation}\label{eq:n_s_ratio}
        \frac{n_s(p)}{n_s(p_c)} = \left(\frac{1 - p}{1 - p_c}\right)^2 \left[\frac{p}{p_c} \left(\frac{1 - p}{1 - p_c}\right)^{z - 2}\right]^s \approx \left[1 - a(p_c - p)^{1/\sigma}\right]^s \propto \exp(-cs),
\end{equation}

\noindent where

\begin{equation*}
    a = \frac{1}{2p_c^2(1 - p_c)},~\sigma = 1/2,~\mathrm{and}~ c = -\ln(1 - a(p_c - p)^{1/\sigma}) \propto (p_c - p)^{1/\sigma}.
\end{equation*}

Above or below the percolation threshold, the cluster size distribution decays exponentially with a parameter proportional to $(p_c - p)^2$. At the threshold, the system has no finite length scale (Eq.~\ref{eq:corr_len_l}), which provides one motivation for assuming that the cluster size distribution at criticality has the form of a power law, $n_s(p_c) \propto s^{-\tau}$. We therefore have $n_s(p) \propto s^{-\tau} e^{-cs}$.

To solve for $\tau$, we can use $n_s$ to derive $S$ another way. The probability that an arbitrary site belongs to a cluster of size $s$ is $s n_s$, because it could be any of the $s$ cluster sites. Since each occupied site belongs to exactly one cluster, we have $\sum s n_s = p$. The mean cluster size is then

\begin{equation}\label{eq:mean_s_ns}
    \begin{split}
        S &= \textstyle\sum s^2 n_s/\textstyle\sum s n_s\\
        &= (1/p)\textstyle\sum s^2 n_s \propto \textstyle\sum s^{2 - \tau} e^{-cs} \approx \textstyle\int s^{2 - \tau} e^{-cs}\\
        &= c^{\tau - 3} \textstyle\int u^{2 - \tau}e^{-u} \mathrm{d}u \propto (p_c - p)^{-(3 - \tau)/\sigma}.
    \end{split}
\end{equation}

Comparing Eq.~\ref{eq:mean_s_l} and Eq.~\ref{eq:mean_s_ns}, we have $1 = (3 - \tau)/\sigma$, giving $\tau = 5/2$. We conclude that finite percolation clusters have a size distribution with power-law exponent $5/2$ and an exponential decay away from criticality.

Finally, we examine the geometry of percolation clusters in terms of Euclidean distance $r$. The radius of a complicated object can be described by its radius of gyration $R_s$,

\begin{equation}
    R_s^2 = \frac{1}{s}\sum_{i = 1}^s |r_i - r_0|^2,
\end{equation}

\noindent where $r_0 = \sum_{i=1}^s r_i / s$ is the center of mass. In fact, $R_s$ is related to the expected distance between any two cluster sites,

\begin{equation}\label{eq:gyration_radius_ij}
    R_s^2 = \frac{1}{2}\frac{1}{s^2}\sum_{ij}|r_i - r_j|^2.
\end{equation}

In addition, percolation clusters are fractal objects. As such, the scaling of their size with linear dimension is naturally characterized by a fractal dimension $D$,

\begin{equation}\label{eq:fractal_dimension}
    R_s = s^{1/D}.
\end{equation}

Next, analogously to Eq.~\ref{eq:corr_fn_l_def} and Eq.~\ref{eq:corr_len_l}, we define the correlation function in Euclidean space $g(r)$ and the corresponding correlation length,

\begin{equation}\label{eq:corr_len_r}
    \xi_r^2 = \frac{\sum_r r^2 g(r)}{\sum_r g(r)}.
\end{equation}

To get an indirect expression for $g(r)$, we recompute $S$ analogously to Eq.~\ref{eq:mean_s_l} and equate it to Eq.~\ref{eq:mean_s_ns}, obtaining the relation

\begin{equation}\label{eq:mean_s_r}
    \sum_r g(r) = \frac{1}{p}\sum_s s^2 n_s.
\end{equation}

The Bethe lattice model doesn't intrinsically define a meaningful notion of Euclidean distance. However, any path along a cluster embedded in a very high-dimensional lattice behaves like a random walk, so that $r^2 \propto l$. Therefore,

\begin{equation}\label{eq:xi_r_to_xi_l}
    \xi_r^2 \propto \xi_l \propto (p_c - p)^{-2\nu},~\nu=1/2.
\end{equation}

Substituting Eq.~\ref{eq:gyration_radius_ij}, Eq.~\ref{eq:fractal_dimension}, and Eq.~\ref{eq:mean_s_r} into Eq.~\ref{eq:corr_len_r} yields

\begin{equation}
    \xi_r^2 = \frac{2\sum_s s^{2 + 2/D} n_s}{\sum_s s^2 n_s}.
\end{equation}

The sums can be computed by generalizing the solution to Eq.~\ref{eq:mean_s_ns} for general moments $M_k$,

\begin{equation}
    M_k = \sum_s s^{k - \tau} e^{-cs} \approx c^{-(k + 1 - \tau)}\int u^{k - \tau}e^{-u}\mathrm{d}u \propto |p_c - p|^{-(k + 1 - \tau)/\sigma}.
\end{equation}

It follows that

\begin{equation}\label{eq:xi_r_soln}
    \xi_r^2 \propto |p_c - p|^{-2/D\sigma}.
\end{equation}

Finally, equating the exponents in Eq.~\ref{eq:xi_r_to_xi_l} and Eq.~\ref{eq:xi_r_soln} yields $D = 1/\sigma\nu = 4$. Ergo, percolation clusters embedded in a high-dimensional lattice have fractal dimension 4.

\end{document}